\documentclass[10pt,twocolumn,letterpaper]{article}

\usepackage{cvpr}
\usepackage{times}
\usepackage{epsfig}
\usepackage{amsmath}
\usepackage{amssymb}
\usepackage{tabulary}
\usepackage{multirow}
\usepackage[table]{xcolor}
\usepackage{epstopdf}
\usepackage[T1]{fontenc}
\usepackage[utf8]{inputenc}
\usepackage{authblk}
\usepackage{url}
\usepackage{graphicx}

\definecolor{lb}{rgb}{0.4, 0.6, 0.8}
\definecolor{br}{rgb}{0.65, 0.16, 0.16}
\definecolor{gc}{rgb}{0.0, 0.5, 0.0}

\usepackage{caption}
\newcommand{\kdtree}{$k$d-tree~}
\newcommand{\ptdn}{\mbox{$\mathbf{X}$}}
\newcommand{\pd}{\mbox{$\mathbf{x}$}}       

\newcommand{\nspacet}{\mbox{$\mathbf{\Psi}$}}
\newcommand{\nspace}{\mbox{$\mathbf{\psi}$}}

\newcommand{\fdup}{\mbox{$\mathcal{J}_{up}$}}
\newcommand{\fdlw}{\mbox{$\mathcal{J}_{lw}$}}
\newcommand{\fdlt}{\mbox{$\mathcal{J}_{lt}$}}
\newcommand{\fdrt}{\mbox{$\mathcal{J}_{rt}$}}
\newcommand{\fdps}{\mbox{$\mathcal{J}_{all}$}}

\newcommand{\rom}[1]{\textup{\uppercase\expandafter{\romannumeral#1}}
}

\DeclareMathOperator*{\argmin}{arg\,min}
\DeclareMathOperator*{\argmax}{arg\,max}

\usepackage{color}
\definecolor{darkblue}{rgb}{0.4,0.4,.8}

\usepackage[pagebackref=true,breaklinks=true,letterpaper=true,colorlinks,bookmarks=false]{hyperref}

\cvprfinalcopy 

\def\cvprPaperID{1374} 

\ifcvprfinal\fi
\begin{document}

\title{A Dual-Source Approach for 3D Pose Estimation from a Single Image\vspace{-7mm}}
\newcommand*\samethanks[1][\value{footnote}]{\footnotemark[#1]}

\author[1]{Hashim Yasin\thanks{authors contributed equally}}
\author[2]{Umar Iqbal\samethanks}
\author[3]{Bj{\"o}rn Kr{\"u}ger}
\author[1]{Andreas Weber}
\author[2]{Juergen Gall\vspace{-1mm}}
\affil[1]{Multimedia, Simulation, Virtual Reality Group, University of Bonn, Germany}
\affil[2]{Computer Vision Group, University of Bonn, Germany}
\affil[3]{Gokhale Method Institute, Stanford, USA\vspace{-7mm}}

\maketitle


\begin{abstract}
One major challenge for 3D pose estimation from a single RGB image is the acquisition of sufficient training data. In particular, collecting large amounts of training data that contain unconstrained images and are annotated with accurate 3D poses is infeasible. We therefore propose to use two independent training sources. The first source consists of images with annotated 2D poses and the second source consists of accurate 3D motion capture data. To integrate both sources, we propose a dual-source approach that combines 2D pose estimation with efficient and robust 3D pose retrieval. In our experiments, we show that our approach achieves state-of-the-art results and is even competitive when the skeleton structure of the two sources differ substantially. 
\end{abstract}

\vspace{-4mm}
\section{Introduction}
Human 3D pose estimation from a single RGB image is a very challenging task. One approach to solve this task is to collect  training data, where each image is annotated with the 3D pose. A regression model, for instance, can then be learned to predict the 3D pose from the image~\cite{Bo-2010,Ilya_2014,ics-cvpr14,Agarwal:2006, bo2008fast,LiC14, Sijin2015iccv}. In contrast to 2D pose estimation, however, acquiring accurate 3D poses for an image is very elaborate. Popular datasets like HumanEva~\cite{Sigal_2010} or Human3.6M~\cite{h36m_pami} synchronized cameras with a commercial marker-based system to obtain 3D poses for images. This requires a very expensive hardware setup and the requirements for marker-based system like studio environment and attached markers prevent the capturing of realistic images.

\begin{figure*}[t]
\begin{center}
 i\includegraphics[width=1\linewidth]{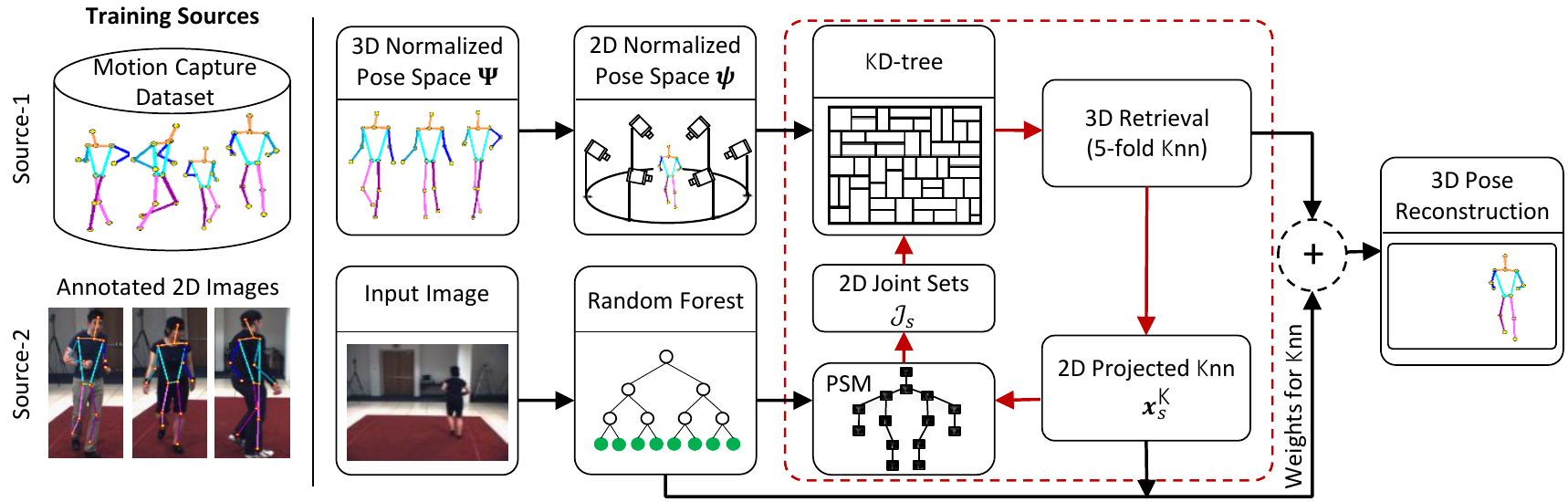}
\end{center}
\vspace{-2mm}	
\caption{
{\bf Overview.} Our approach relies on two training sources. The first source is a motion capture database that contains only 3D poses. The second source is an image database with annotated 2D poses. The motion capture data is processed by pose normalization and projecting the poses to 2D using several virtual cameras. This gives many 3D-2D pairs where the 2D poses serve as features. The image data is used to learn a pictorial structure model (PSM) for 2D pose estimation where the unaries are learned by a random forest. Given a test image, the PSM predicts the 2D pose which is then used to retrieve the normalized nearest 3D poses. The final 3D pose is then estimated by minimizing the projection error under the constraint that the solution is close to the retrieved poses, which are weighted by the unaries of the PSM. The steps (\emph{red arrows}) in the dashed box can be iterated by updating the binaries of the PSM using the retrieved poses and updating the 2D pose.
}
\vspace{-2mm}
\label{fig:sysflow}
\end{figure*}

Instead of training a model on pairs consisting of an image and a 3D pose, we propose an approach that is able to incorporate 2D and 3D information from two different training sources.
The first source consists of images with annotated 2D pose. Since 2D poses in images can be manually annotated, they do not impose any constraints regarding the environment from where the images are taken. Indeed any image from the Internet can be annotated and used. The second source is accurate 3D motion capture data captured in a lab, \eg, as in the CMU motion capture dataset~\cite{cmu_mocap} or the Human3.6M dataset \cite{h36m_pami}. We consider both sources as independent, \ie, we do not know the 3D pose for any training image. To integrate both sources, we propose a dual-source approach as illustrated in Fig.~\ref{fig:sysflow}. To this end, we first convert the motion capture data into a normalized 2D pose space, and separately learn a regressor for 2D pose estimation from the image data. During inference, we estimate the 2D pose and retrieve the nearest 3D poses using an approach that is robust to 2D pose estimation errors.
We then jointly estimate a mapping from the 3D pose space to the image, identify wrongly estimated 2D joints, and estimate the 3D pose. During this process, the 2D pose can also be refined and the approach can be iterated to update the estimated 3D and 2D pose. We evaluate our approach on two popular datasets for 3D pose estimation. On both datasets, our approach achieves state-of-the-art results and we provide a thorough evaluation of the approach. In particular, we analyze the impact of differences of the skeleton structure between the two training sources, the impact of the accuracy of the used 2D pose estimator, and the impact of the similarity of the training and test poses.

\section{Related Work}
A common approach for 3D human pose estimation is to utilize multiple images captured by synchronized cameras \cite{belagiannis20143d, sigal2012loose, yao2012coupled}.
The requirement of a multi-camera system in a controlled environment, however, limits the applicability of these methods. Since 3D human pose estimation from a single image is very difficult due to missing depth information, depth cameras have been utilized for human pose estimation~\cite{Baak:2011,Shotton:2011,Grest:2005}.
However, current depth sensors are limited to indoor environments and cannot be used in unconstrained scenarios.
Earlier approaches for monocular 3D human pose estimation \cite{Bo-2010, AgarwalT-2004,Sminchisescu-2005,Agarwal:2006, bo2008fast, mori2006recovering} utilize discriminative methods to learn a mapping from local image features (\eg HOG, SIFT, \etc ) to 3D human pose or use a CNN~\cite{LiC14, Sijin2015iccv}. Since local features are sensitive to noise, these methods often assume that the location and scale of the human is given, \eg, in the form of an accurate bounding box or silhouette.
While the approach~\cite{ics-cvpr14} still relies on the known silhouette of the human body, it partially overcomes the limitations of local image features by segmenting the body parts and using a second order hierarchical pooling process to obtain robust descriptors. Instead of predicting poses with a low 3D joint localization error, an approach for retrieving semantic meaningful poses is proposed in~\cite{Pons-Moll_2014_CVPR}.  

The 3D pictorial structure model (PSM) proposed in \cite{Ilya_2014} combines generative and discriminative methods. Regression forests are trained to estimate the probabilities of 3D joint locations and the final 3D pose is inferred by the PSM. Since inference is performed in 3D, the bounding volume of the 3D pose space needs to be known and the inference requires a few minutes per frame.

Besides of a-priori knowledge about bounding volumes, bounding boxes or silhouettes, these approaches require sufficient training images with annotated 3D poses. Since such training data is very difficult to acquire, we propose a dual-source approach that does not require training images with 3D annotations, but exploits existing motion capture datasets to estimate the 3D human pose.

Estimating 3D human pose from a given 2D pose by exploiting motion capture data has been addressed in a few works~\cite{SimoSerraCVPR2012,Ramakrishna_2012,yasin-2013,SimoSerraCVPR2013,Wang_2014_CVPR}. In~\cite{yasin-2013}, the 2D pose is manually annotated in the first frame and tracked in a video. A nearest neighbor search is then performed to retrieve the closest 3D poses.
In~\cite{Ramakrishna_2012} a sparse representation of 3D human pose is constructed from a MoCap dataset and fitted to manually annotated 2D joint locations. The approach has been extended in \cite{Wang_2014_CVPR} to handle poses from an off-the-shelf 2D pose estimator~\cite{YiYang-2011}. The same 2D pose estimator is also used in \cite{SimoSerraCVPR2012,SimoSerraCVPR2013} to constrain the search space of 3D poses. In \cite{SimoSerraCVPR2012} an evolutionary algorithm is used to sample poses from the pose space that correspond to the estimated 2D joint positions. This set is then exhaustively evaluated according to some anthropometric constraints. The approach is extended in~\cite{SimoSerraCVPR2013} such that the 2D pose estimation and 3D pose estimation are iterated. In contrast to~\cite{Ramakrishna_2012,Wang_2014_CVPR,SimoSerraCVPR2012}, \cite{SimoSerraCVPR2013} deals with 2D pose estimation errors. Our approach also estimates 2D and 3D pose but it is faster and more accurate than the sampling based approach~\cite{SimoSerraCVPR2013}.

Action specific priors learned from the MoCap data have also been proposed for 3D pose tracking ~\cite{UrtasunFF06,420}. These approaches, however, are more constrained by assuming that the type of motion is known in advance. 

\vspace{-2.1mm}
\section{Overview}\label{sec:overview}
In this work, we aim to predict the 3D pose from an RGB image. Since acquiring 3D pose data in natural environments is impractical and annotating 2D images with 3D pose data is infeasible, we do not assume that our training data consists of images annotated with 3D pose. Instead, we propose an approach that utilizes two independent sources of training data. The first source consists of motion capture data, which is publically available in large quantities and that can be recorded in controlled environments. The second source consists of images with annotated 2D poses, which is also available and can be easily provided by humans. Since we do not assume that we know any relations between the sources except that the motion capture data includes the poses we are interested in, we preprocess the sources first independently as illustrated in Fig.~\ref{fig:sysflow}. From the image data, we learn a pictorial structure model (PSM) to predict 2D poses from images. This will be discussed in Section~\ref{sec:posdet}. The motion capture data is prepared to efficiently retrieve 3D poses that could correspond to a 2D pose. This part is described in Section \ref{sec:similarity}. We will show that the retrieved poses are insufficient for estimating the 3D pose. Instead, we estimate the pose by minimizing the projection error under the constraint that the solution is close to the retrieved poses (Section~\ref{sec:posrec}).
In addition, the retrieved poses can be used to update the PSM and the process can be iterated (Section \ref{sec:iter}). In our experiments, we show that we achieve very good results for 3D pose estimation with only one or two iterations.

The models for 2D pose estimation and the source code for 3D pose estimation are publicly available. \footnote{\url{http://pages.iai.uni-bonn.de/iqbal_umar/ds3dpose/}}

\section{2D Pose Estimation}\label{sec:posdet}
In this work, we adopt a PSM that represents the 2D body pose \pd~with a graph $\mathcal{G}=(\mathcal{J}, \mathcal{L})$, where each vertex corresponds to 2D coordinates of a particular body joint $i$, and edges correspond to the kinematic constraints between two joints $i$ and $j$. We assume that the graph is a tree structure which allows efficient inference. Given an image $\mathbf{I}$, the 2D body pose is inferred by maximizing the following posterior distribution,

\begin{equation}
\label{eqt:ps}
P(\bold{x}|\bold{I}) \propto \prod_{i \in \mathcal{J}} \phi_i(x_{i}|\bold{I}) \prod_{(i,j) \in \mathcal{L}} \phi_{i,j}(x_i, x_j),
\end{equation}
where the unary potentials $\phi_i(x_{i}|\bold{I})$ correspond to joint templates and define the probability of the $i^{th}$ joint at location $x_i$. The binary potentials $\phi_{i,j}(x_i, x_j)$ define the deformation cost of joint $i$ from its parent joint $j$. 

The unary potentials in \eqref{eqt:ps} can be modeled by any discriminative model, \eg, SVM in \cite{YiYang-2011} or random forests in \cite{dantone_tpami2014}. In this work, we choose random forest based joint regressors. We train a separate joint regressor for each body joint. Following \cite{dantone_tpami2014}, we model binary potentials for each joint $i$ as a Gaussian mixture model with respect to its parent $j$. We obtain the relative joint offsets between two adjacent joints in the tree structure and cluster them into $c=1, \dots ,C$ clusters using k-means clustering. The offsets in each cluster are then modeled with a weighted Gaussian distribution as,
\begin{equation}\label{eq:binary}
\gamma_{ij}^{c} \exp\left(-\dfrac{1}{2}\left(d_{ij}-\mu_{ij}^{c}\right)^T\left(\Sigma_{ij}^{c}\right)^{-1}\left(d_{ij}-\mu_{ij}^{c}\right)\right)
\end{equation}
with mean $\mu_{ij}^{c}$, covariance $\Sigma_{ij}^{c}$ and  $d_{ij} = (x_i{-}x_j)$. The weights $\gamma_{ij}^{c}$ are set according to the cluster frequency $p(c|i,j)^\alpha$ with a normalization constant $\alpha=0.1$ \cite{dantone_tpami2014}.

\section{3D Pose Estimation}
While the PSM for 2D pose estimation is trained on the images with 2D pose annotations as shown in Fig.~\ref{fig:sysflow}, we now describe an approach that makes use of a second dataset with 3D poses in order to predict the 3D pose from an image. Since the two sources are independent, we first have to establish relations between 2D poses and 3D poses. This is achieved by using an estimated 2D pose as query for 3D pose retrieval (Section \ref{sec:similarity}). The retrieved poses, however, contain many wrong poses due to errors in 2D pose estimation, 2D-3D ambiguities and differences of the skeletons in the two training sources. It is therefore necessary to fit the 3D poses to the 2D observations. This will be described in Section~\ref{sec:posrec}.
\subsection{3D Pose Retrieval}\label{sec:similarity}
\begin{figure}
\begin{center}
\includegraphics[width=.9\linewidth]{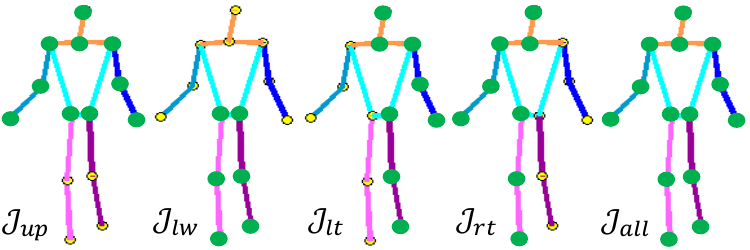}
\end{center}
   \caption{Different joint sets. \fdup~is based on upper body joints, \fdlw~lower body joints, \fdlt~left body joints, \fdrt~right body joints and \fdps~is composed of all body joints. The selected joints are indicated by the large green circles. 
   }
\label{fig:features}
\end{figure}
In order to efficiently retrieve 3D poses for a 2D pose query, we preprocess the motion capture data. We first normalize the poses by discarding orientation and translation information from the poses in our motion capture database. We denote a 3D normalized pose with \ptdn~and the 3D normalized pose space with \nspacet. As in \cite{yasin-2013}, we project the normalized poses $\ptdn \in \nspacet$ to 2D using orthographic projection. We use 144 virtual camera views with azimuth angles spanning 360 degrees and elevation angles in the range of 0 and 90 degree. Both angles are uniformly sampled with step size of 15 degree. We further normalize the projected 2D poses by scaling them such that the y-coordinates of the joints are within the range of $[-1, 1]$. The normalized 2D pose space is denoted by \nspace\ and does not depend on a specific camera model or coordinate system. This step is illustrated in Fig.~\ref{fig:sysflow}. After a 2D pose is estimated by the approach described in Section~\ref{sec:posdet}, we first normalize it according to \nspace, \ie, we translate and scale the pose such that the y-coordinates of the joints are within the range of $[-1, 1]$,  then use it to retrieve 3D poses. The distance between two normalized 2D poses is given by the average Euclidean distance of the joint positions. The $\mathsf{K}$-nearest neighbours in \nspace\ are efficiently retrieved by a \kdtree~\cite{krueger-2010}. The retrieved normalized 3D poses are the corresponding poses in \nspacet. An incorrect 2D pose estimation or even an imprecise estimation of a single joint position, however, can effect the accuracy of the 3D pose retrieval and consequently the 3D pose estimation. We therefore propose to use several 2D joint sets for pose retrieval where each joint set contains a different subset of all joints. The joint sets are shown in Fig.~\ref{fig:features}. While \fdps\ contains all joints, the other sets \fdup, \fdlw, \fdlt\ and \fdrt\ contain only the joints of the upper body, lower body, left hand side and right hand side of the body, respectively. In this way we are able to compensate for 2D pose estimation errors, if at least one of our joint sets does not depend on the wrongly estimated 2D joints.

\subsection{3D Pose Estimation}\label{sec:posrec}
In order to obtain the 3D pose $\mathbf{X}$, we have to estimate the unknown projection $\mathcal{M}$ from the normalized pose space \nspacet\ to the image and infer which joint set $\mathcal{J}_s$ explains the image data best. To this end, we minimize the energy
\begin{equation}
E(\mathbf{X},\mathcal{M},s) = \omega_{p}E_{p}(\mathbf{X},\mathcal{M},s) + \omega_{r}E_{r}(\mathbf{X},s) + \omega_{a}E_{a}(\mathbf{X},s)
\label{eq:energyMin}
\end{equation}
consisting of the three weighted terms $E_{p}$, $E_{r}$ and $E_{a}$.

The first term $E_{p}(\mathbf{X},\mathcal{M},s)$ measures the projection error of the 3D pose $\mathbf{X}$ and the projection $\mathcal{M}$:
\begin{equation}\label{eq:energyControl}
E_{p}(\mathbf{X},\mathcal{M},s) = \left( \sum_{i \in \mathcal{J}_{s}} \| \mathcal{M}\left(X_{i}\right) - x_i\|^2 \right)^{\frac{1}{4}},
\end{equation}
where $x_i$ is the joint position of the predicted 2D pose and $X_{i}$ is the 3D joint position of the unknown 3D pose. The parameter $s$ defines the set of valid 2D joint estimates and the error is only computed for the joints of the corresponding joint set $\mathcal{J}_{s}$.

The second term enforces that the pose $\mathbf{X}$ is close to the retrieved 3D poses $\mathbf{X}^{\mathsf{k}}_{s}$ for a joint set $\mathcal{J}_{s}$:
\begin{equation}
E_{r}(\mathbf{X},s) = \sum_{\mathsf{k}} w_{\mathsf{k},s}
 \left( \sum_{i \in \mathcal{J}_{all}} \| X^{\mathsf{k}}_{s,i} - X_i\|^2 \right)^{\frac{1}{4}}.
\label{eq:energyMotion}
\end{equation}
In contrast to \eqref{eq:energyControl}, the error is computed over all joints but the set of nearest neighbors depends on $s$. In our experiments, we will show that an additional weighting of the nearest neighbors by $w_{\mathsf{k},s}$ improves the 3D pose estimation accuracy.

Although the term $E_{r}(\mathbf{X},s)$ penalizes already deviations from the retrieved poses and therefore enforces implicitly anthropometric constraints, we found it useful to add an additional term that enforces anthropometric constraints on the limbs:
\begin{equation}
E_{a}(\mathbf{X},s)  = \sum_{\mathsf{k}} w_{\mathsf{k},s}
 \left( \sum_{(i,j) \in \mathcal{L}} \left( L^{\mathsf{k}}_{s,i,j} - L_{i,j} \right)^2 \right)^{\frac{1}{4}},
\label{eq:energyLimb}
\end{equation}
where $L_{i,j}$ denotes the limb length between two joints.

Minimizing the energy $E(\mathbf{X},\mathcal{M},s)$ \eqref{eq:energyMin} over the discrete variable $s$ and the continuous parameters $\mathbf{X}$ and $\mathcal{M}$ would be expensive. We therefore propose to obtain an approximate solution where we estimate the projection $\mathcal{M}$ first. For the projection, we assume that the intrinsic parameters are given and only estimate the global orientation and translation.
The projection $\hat{\mathcal{M}}_{s}$ is estimated for each joint set $\mathcal{J}_s$ with $s \in \{ up, lw, lt, rt, all \}$ by minimizing
\begin{equation}
\hat{\mathcal{M}}_{s} = \argmin_{\mathcal{M}} \left\{ \sum_{\mathsf{k}=1}^{\mathsf{K}} E_{p}(\mathbf{X}^{\mathsf{k}}_{s},\mathcal{M},s) \right\}
\label{eq:projerr1}
\end{equation}
using non-linear gradient optimization. Given the estimated projections $\hat{\mathcal{M}}_{s}$ for each joint set, we then optimize over the discrete variable $s$:
\begin{equation}
\hat{s} = \argmin_{s \in \{ up, lw, lt, rt, all \}} \left\{ \sum_{\mathsf{k}=1}^{\mathsf{K}} E(\mathbf{X}^{\mathsf{k}}_{s},\hat{\mathcal{M}}_s,s) \right\}.
\label{eq:energyS}
\end{equation}
As a result, we obtain $\hat{s}$ and $\hat{\mathcal{M}}=\hat{\mathcal{M}}_{\hat{s}}$ and finally minimize
\begin{equation}
\hat{\mathbf{X}} = \argmin_{\mathbf{X}} \left\{ E(\mathbf{X},\hat{\mathcal{M}},\hat{s}) \right\}
\label{eq:energyMin2}
\end{equation}
to obtain the 3D pose.

\paragraph{Implementation details} Instead of obtaining $\hat{s}$ by minimizing~\eqref{eq:energyS}, $\hat{s}$ can also be estimated by maximizing the posterior distribution for the 2D pose~\eqref{eqt:ps}.
To this end, we project all retrieved 3D poses to the image by
\begin{equation}
x^{\mathsf{k}}_{s,i} =  \hat{\mathcal{M}}_s \left( X^{\mathsf{k}}_{s,i} \right).
\label{eq:projerr2}
\end{equation}
The binary potentials $\phi_{i,j}(x_i, x_j|\mathbf{X}_s)$, which are mixture of Gaussians, are then computed from the projected full body poses for each set and $\hat{s}$ is inferred by the maximum posterior probability:
\begin{equation}
\label{eqt:best_segments_pose}
(\mathbf{\hat{x}},\hat{s}) = \argmax_{\mathbf{x},s} \left\{ \prod_{i \in \mathcal{J}} \phi_i(x_{i}|\bold{I}) \prod_{i,j \in \mathcal{L}} \phi_{i,j}(x_i, x_j|\mathbf{X}_s) \right\}.
\end{equation}
Finally, the refined 2D pose $\mathbf{\hat{x}}$ is used to compute the projection error $E_{p}(\mathbf{X},\hat{\mathcal{M}},\hat{s})$ in \eqref{eq:energyMin2}. 

In addition, we weight the nearest neighbors by
\begin{equation}
\mathsf{w}_{\mathsf{k},s} = \sum_{i \in \mathcal{J}} \phi_i(x^{\mathsf{k}}_{s,i}|\bold{I}),
\label{eq:wt}
\end{equation}
to keep only the $\mathsf{K}_w$ poses with the highest weights, and normalize them by
\begin{equation}
w_{\mathsf{k},s} = \frac{\mathsf{w}_{\mathsf{k},s} - \min_{\mathsf{k}'}(\mathsf{w}_{\mathsf{k}',s})}{ \max_{\mathsf{k}'}(\mathsf{w}_{\mathsf{k}',s}) - \min_{\mathsf{k}'}(\mathsf{w}_{\mathsf{k}',s})}.
\label{eq:wtn}
\end{equation}
The dimensionality of $\mathbf{X}$ can be reduced by applying PCA to the weighted poses.
We thoroughly evaluate the impact of the implementation details in Section~\ref{sec:par}.

\subsection{Iterative Approach}\label{sec:iter}
The approach can be iterated by using the refined 2D pose $\mathbf{\hat{x}}$~\eqref{eqt:best_segments_pose} as query for 3D pose retrieval (Section \ref{sec:similarity}) as illustrated in Fig.~\ref{fig:sysflow}. Having more than one iteration is not very expensive since many terms like the unaries need to be computed only once and the optimization of \eqref{eq:projerr1} can be initialized by the results of the previous iteration. The final pose estimation \eqref{eq:energyMin2} also needs to be computed only once after the last iteration. In our experiments, we show that two iterations are sufficient.
%
%
%
\section{Experiments}\label{sec:exp}
We evaluate the proposed approach on two publicly available datasets, namely HumanEva-I~\cite{Sigal_2010} and Human3.6M~\cite{h36m_pami}.
Both datasets provide accurate 3D poses for each image and camera parameters. For both datasets, we use a skeleton consisting of 14 joints, namely head, neck, ankles, knees, hips, wrists, elbows and shoulders.
For evaluation, we use the \textit{3D pose error} as defined in~\cite{SimoSerraCVPR2012}. The error measures the accuracy of the relative pose up to a rigid transformation. To this end, the estimated skeleton is aligned to the ground-truth skeleton by a rigid transformation and the average 3D Euclidean joint error after alignment is measured. In addition, we use the CMU motion capture dataset~\cite{cmu_mocap} as training source.

%
%
\vspace{-1mm}
\subsection{Evaluation on HumanEva-I Dataset}
We follow the same protocol as described in~\cite{SimoSerraCVPR2013,Ilya_2014} and use the provided training data to train our approach while using the validation data as test set. As in~\cite{SimoSerraCVPR2013,Ilya_2014}, we report our results on every $5^{th}$ frame of the sequences \textit{walking} (A1) and \textit{jogging} (A2) for all three subjects (S1, S2, S3) and camera C1. For 2D pose estimation, we train regression forests and PSMs for each activity as described in~\cite{dantone_tpami2014}. The regression forests for each joint consists of 8 trees, each trained on 700 randomly selected training images from a particular activity. While we use $c = 15$ mixtures per part \eqref{eq:binary} for the initial pose estimation, we found that 5 mixtures are enough for pose refinement (Section~\ref{sec:posrec}) since the retrieved 2D nearest neighbours strongly reduce the variation compared to the entire training data. In our experiments, we consider two sources for the motion capture data, namely HumanEva-I and the CMU motion capture dataset. We first evaluate the parameters of our approach using the entire 49K 3D poses of the HumanEva training set as motion capture data. Although the training data for 2D pose estimation and the 3D pose data are from the same dataset, the sources are separated and it is unknown which 3D pose corresponds to which image.

\vspace{-2.5mm}
\subsubsection{Parameters}\label{sec:par}
\begin{figure}[t]
\begin{center}

\includegraphics[width=.85\linewidth]{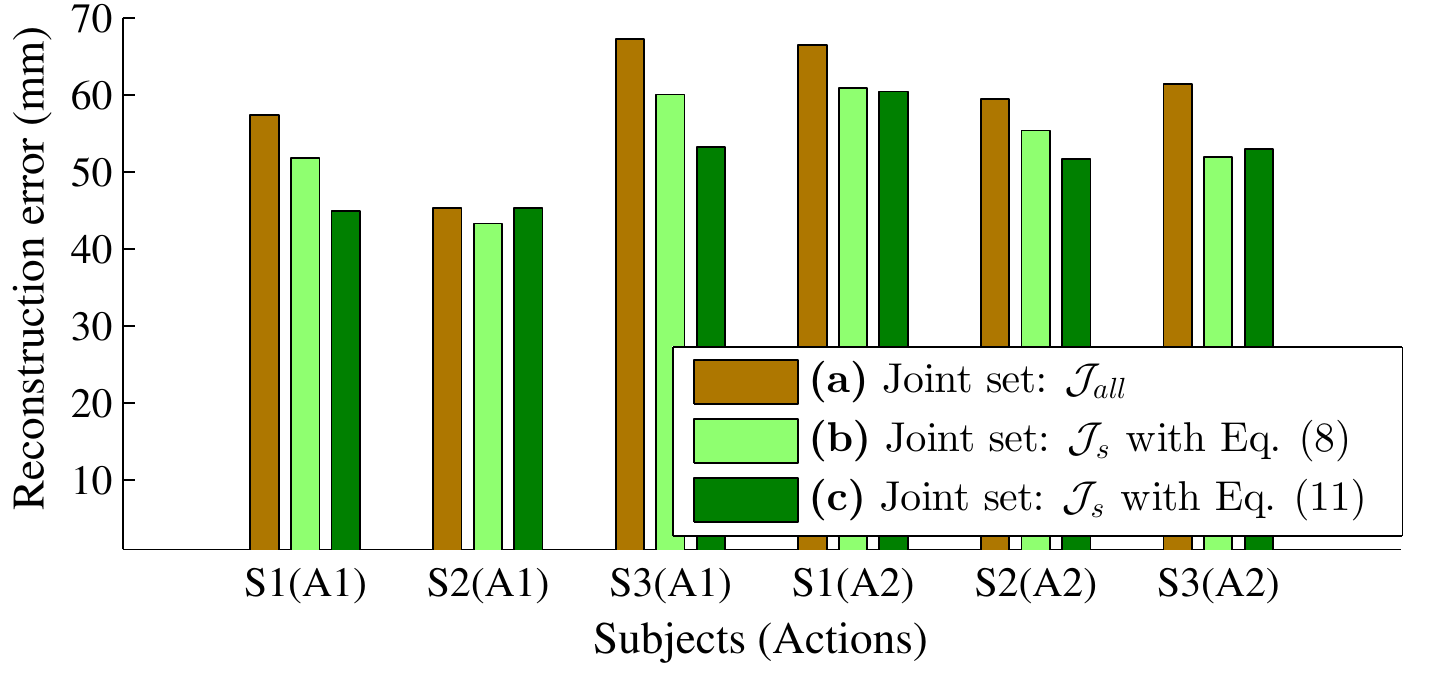}
\end{center}
   \vspace{-2mm}	
   \caption{\textbf{(a)} Using only joint set \fdps. \textbf{(b)} Using all joint sets $\mathcal{J}_{s}$ and estimating $\hat{s}$ using~\eqref{eq:energyS}. \textbf{(c)} All joint sets $\mathcal{J}_{s}$ and estimating $\hat{s}$ using~\eqref{eqt:best_segments_pose}.}
\label{fig:query}
\end{figure}

\begin{figure}[t]
\begin{center}
\includegraphics[width=.94\linewidth]{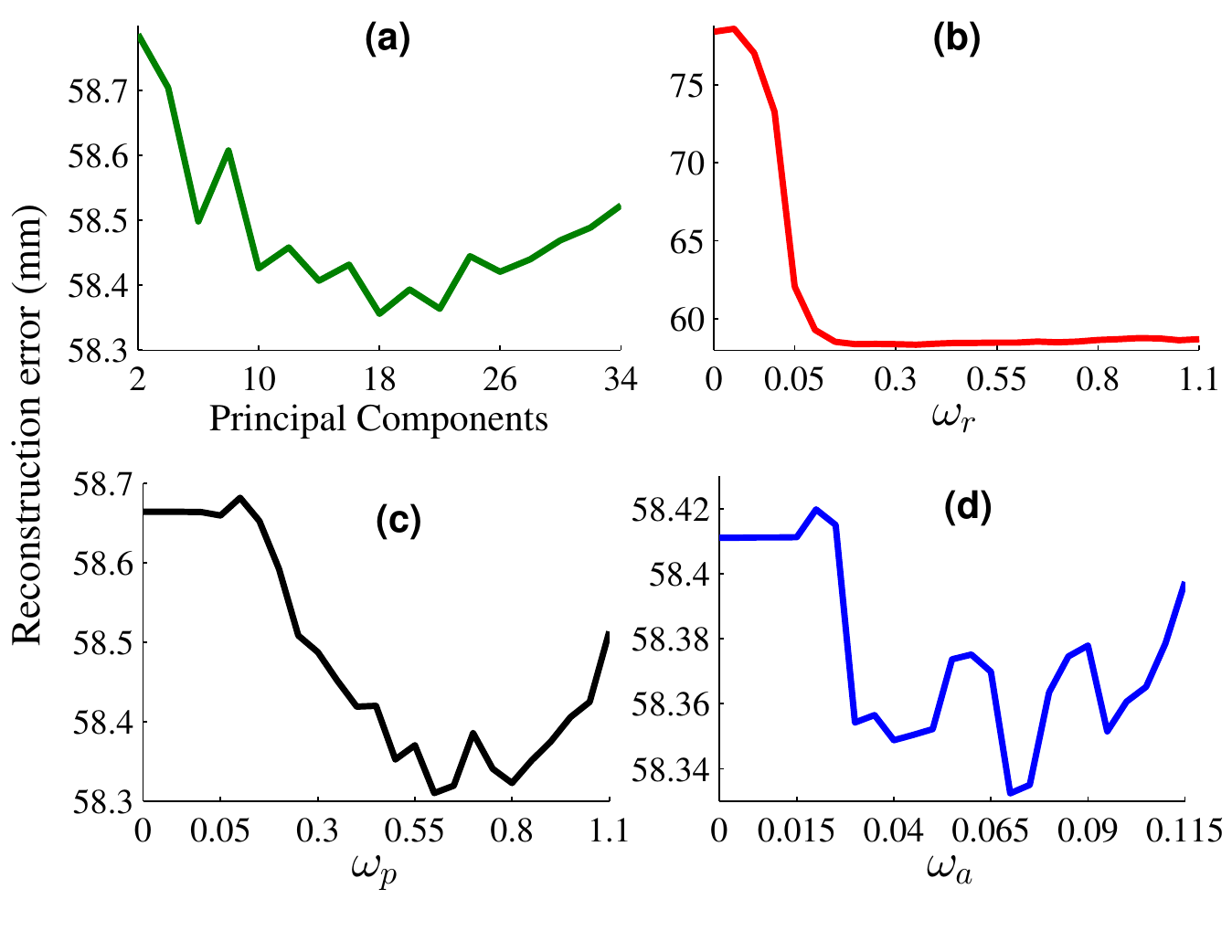}
\end{center}
   \vspace{-5mm}	
   \caption{\textbf{(a)} Impact of the number of principal components. The error is reported for subject S3 with action \textit{jogging} (A2, C1) using the CMU dataset for 3D pose retrieval.  \textbf{(b-d)} Impact of the weights $\omega_r$, $\omega_p$ and $\omega_a$ in \eqref{eq:energyMin}.
 }
\label{fig:energy_pc}
\end{figure}

\paragraph{Joint Sets $\mathcal{J}$.}
For 3D pose retrieval (Section~\ref{sec:similarity}), we use several joint sets $\mathcal{J}_{s}$ with $s \in \{ up, lw, lt, rt, all\}$. For the evaluation, we use only one iteration and $\mathsf{K}=256$ without weighting. The results in Fig.~\ref{fig:query} show the benefit of using several joint sets. Estimating $\hat{s}$ using~\eqref{eqt:best_segments_pose} instead of~\eqref{eq:energyS} also reduces the pose estimation error.
%
\begin{figure}[t]
\begin{center}
\includegraphics[width=.88\linewidth]{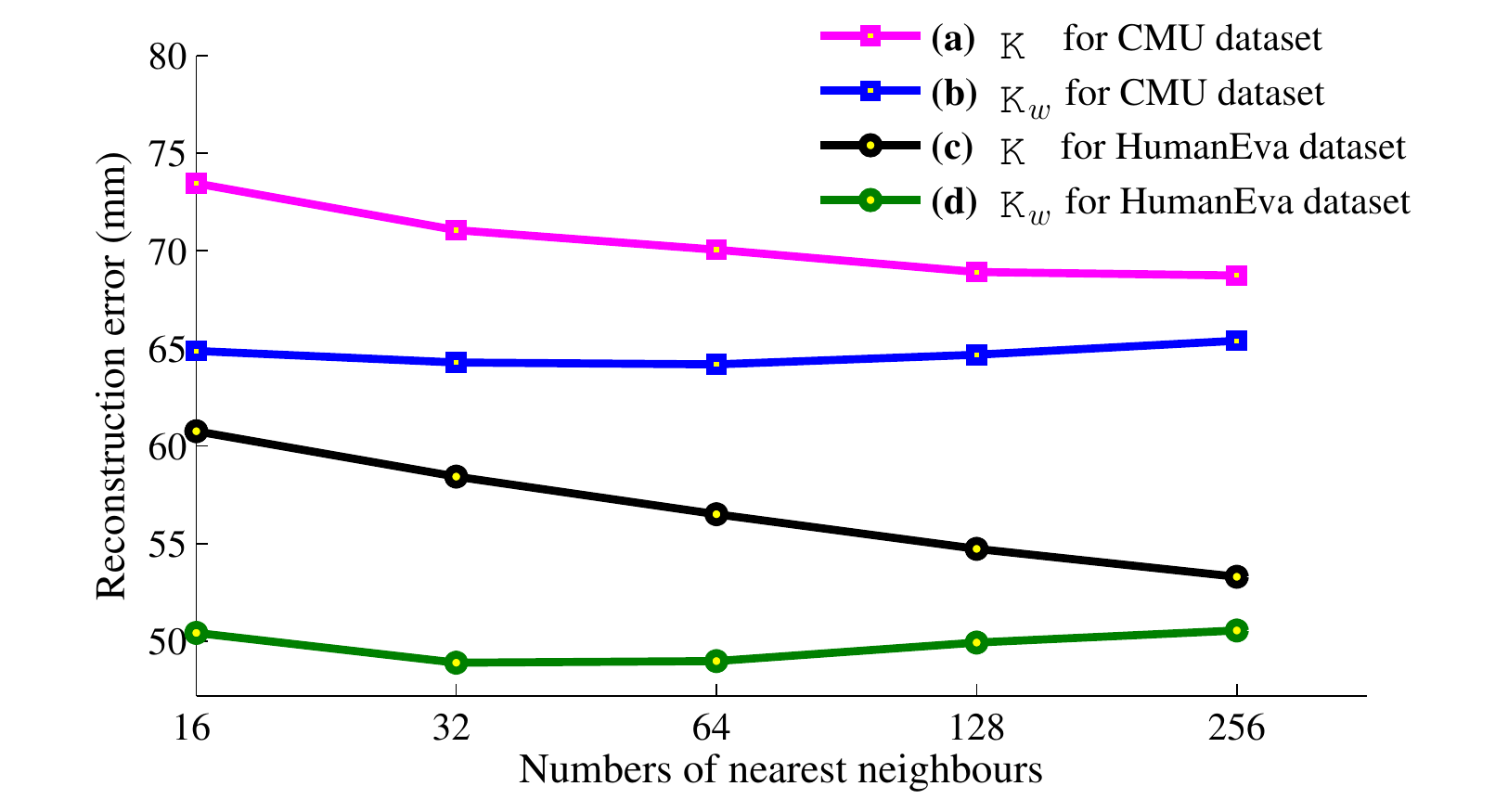}
\end{center}
   \vspace{-2mm}	
   \caption{Impact of the number of nearest neighbours $\mathsf{K}$ and weighting of nearest neighbours $\mathsf{K}_w $. The results are reported for subject S3 with \textit{walking} action (A1, C1) using the CMU dataset \textbf{(a-b)} and HumanEva \textbf{(c-d)} for 3D pose retrieval.}
\label{fig:knn}
\end{figure}
\vspace{-3.5mm}
\paragraph{Nearest Neighbours.}
The impact of weighting the retrieved 3D poses and the number of nearest neighbours is evaluated in Fig.~\ref{fig:knn}. The results show that the weighting reduces the pose estimation error independently of the used motion capture dataset. Without weighting more nearest neighbours are required. If not otherwise specified, we use $\mathsf{K} = 256$ and $\mathsf{K}_w = 64$ for the rest of the paper. 
If the average of the retrieved $\mathsf{K}$ or $\mathsf{K}_w$ poses is used instead of optimizing \eqref{eq:energyMin2}, the errors are $55.7mm$ and $48.9mm$, respectively, as compared to $53.2mm$ and $47.5mm$ by optimizing \eqref{eq:energyMin2}.
PCA can be used to reduce the dimension of $\mathbf{X}$. Fig.~\ref{fig:energy_pc}(a) evaluates the impact of the number of principal components. Good results are achieved for 10-26 components, but the exact number is not critical.
In our experiments, we use 18.
%
\vspace{-3.5mm}
\paragraph{Energy Terms.}
The impact of the weights $\omega_{r}$, $\omega_{p}$ and $\omega_{a}$ in~\eqref{eq:energyMin} is reported in Fig.~\ref{fig:energy_pc}(b-d). Without the term $E_{r}$, the error is very high. This is expected since the projection error $E_{p}$ is evaluated on the joint set $\mathcal{J}_{\hat{s}}$. If $\mathcal{J}_{\hat{s}}$ does not contain all joints, the optimization is not sufficiently constrained without $E_{r}$. Since $E_{r}$ is already weighted by the image consistency of the retrieved poses, $E_{p}$ does not result in a large drop of the error, but refines the 3D pose. The additional anthropometric constraints $E_{a}$ slightly reduce the error in addition. In our experiments, we use $\omega_p = 0.55$, $\omega_r = 0.35$, and $\omega_a = 0.065$.

\begin{figure}[]
\begin{center}
\includegraphics[width=.85\linewidth]{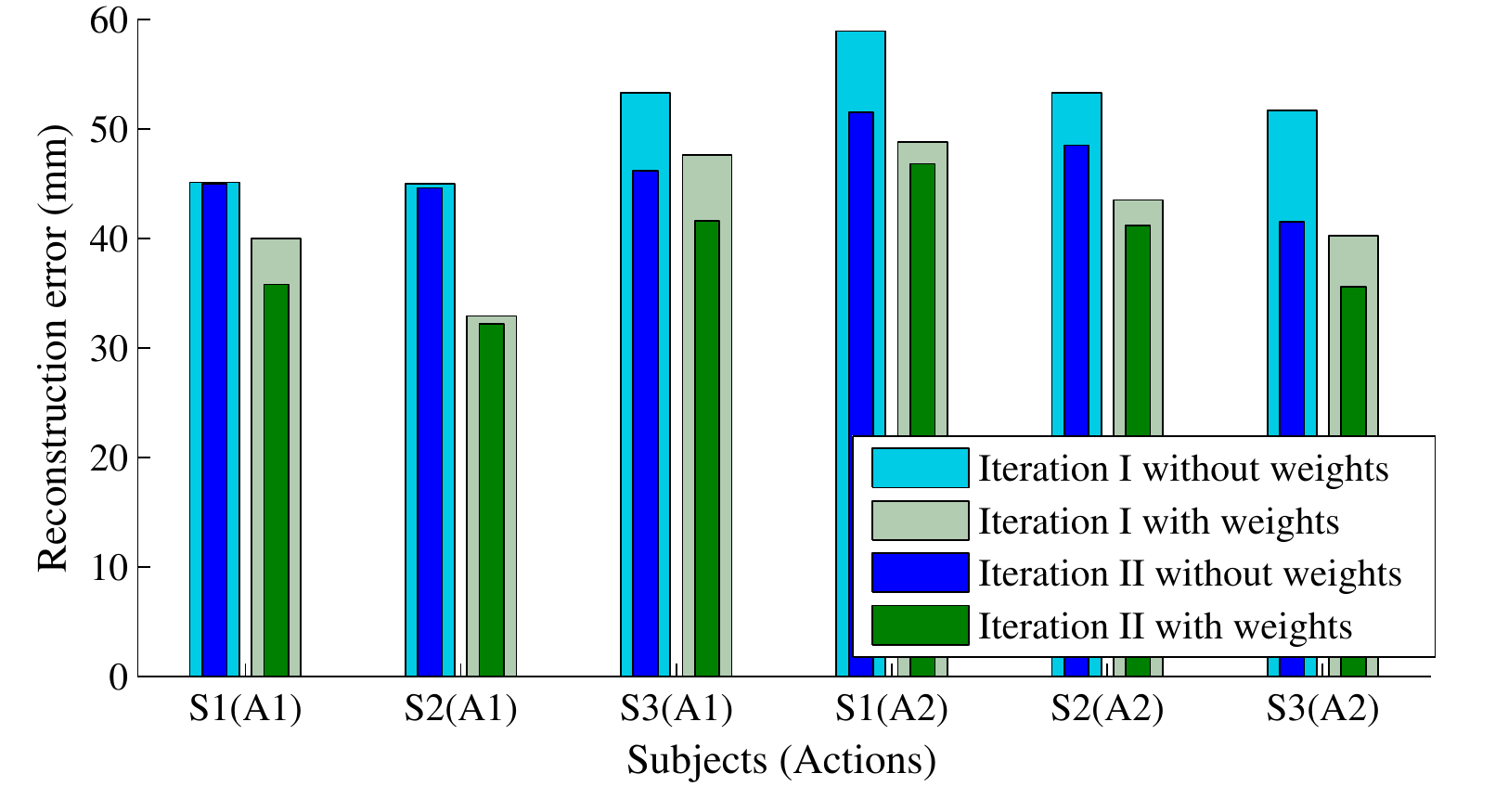}
\end{center}
   \vspace{-5mm}	
   \caption{Impact of the number of iterations and weighting of nearest neighbors.}
\label{fig:wts}
\end{figure}

\vspace{-3.5mm}
\paragraph{Iterations.}
We finally evaluate the benefit of having more than one iteration (Section~\ref{sec:iter}). Fig.~\ref{fig:wts} compares the pose estimation error for one and two iterations. For completeness, the results for nearest neighbours without weighting are included. In both cases, a second iteration decreases the error on nearly all sequences. A third iteration does not reduce the error further.

\begin{table*}[t]
\centering
\scalebox{0.89}{
\begin{tabular}{|l||c|c|c||c|c|c||c|}
\hline
\multicolumn{1}{|c||}{\multirow{2}{*}{Methods} } & \multicolumn{3}{c||}{Walking (A1, C1)}            & \multicolumn{3}{c||}{Jogging (A2, C1)}  & \multicolumn{1}{c|}{\multirow{2}{*}{Average}}   \\ \cline{2-7}

\multicolumn{1}{|c||}{}                  & \multicolumn{1}{c|}{S1} & \multicolumn{1}{c|}{S2} & \multicolumn{1}{c||}{S3} & \multicolumn{1}{c|}{S1} & \multicolumn{1}{c|}{S2} & \multicolumn{1}{c||}{S3} & \\ \hline

Kostrikov \etal \cite{Ilya_2014}           & 44.0 $\pm$ 15.9   & 30.9 $\pm$ 12.0     & 41.7 $\pm$ 14.9     & 57.2 $\pm$ 18.5     & {\bf 35.0} $\pm$ 9.9      & {\bf 33.3} $\pm$ 13.0  & 40.3  $\pm$ 14.0        \\ \hline
Wang \etal \cite{Wang_2014_CVPR}           & 71.9 $\pm$ 19.0   & 75.7 $\pm$ 15.9     & 85.3 $\pm$ 10.3     & 62.6 $\pm$ 10.2     & 77.7 $\pm$ 12.1     & 54.4 $\pm$ 9.0   & 71.3  $\pm$ 12.7         \\ \hline
Radwan \etal \cite{Radwan-2013iccv}     & 75.1 $\pm$ 35.6  & 99.8 $\pm$ 32.6  & 93.8 $\pm$ 19.3 & 79.2 $\pm$ 26.4  & 89.8 $\pm$ 34.2     & 99.4 $\pm$ 35.1 & 89.5 $\pm$ 30.5\\ \hline
Simo-Serra \etal \cite{SimoSerraCVPR2013}  & 65.1 $\pm$ 17.4   & 48.6 $\pm$ 29.0     & 73.5 $\pm$ 21.4     & 74.2 $\pm$ 22.3     & 46.6 $\pm$ 24.7     & 32.2 $\pm$ 17.5  & 56.7  $\pm$ 22.0          \\ \hline
Simo-Serra \etal \cite{SimoSerraCVPR2012}  & 99.6 $\pm$ 42.6   & 108.3 $\pm$ 42.3    & 127.4 $\pm$ 24.0    & 109.2 $\pm$ 41.5    & 93.1 $\pm$ 41.1     & 115.8 $\pm$ 40.6 & 108.9  $\pm$ 38.7         \\ \hline
Bo \etal \cite{Bo-2010} (GT-BB)  & 46.4 $\pm$  20.3   & {\bf 30.3} $\pm$ 10.5     & 64.9 $\pm$ 35.8     & 64.5 $\pm$ 27.5     & 48.0 $\pm$ 17.0     & 38.2 $\pm$ 17.7  & 48.7  $\pm$ 21.5 \\ \hline
Bo \etal \cite{Bo-2010}  (Est-BB)  & 54.8 $\pm$ 40.7   & 36.7 $\pm$ 20.5     & 71.3 $\pm$ 39.8     & 74.2 $\pm$ 47.1     & 51.3 $\pm$ 18.1     & 48.9 $\pm$ 34.2  & 56.2  $\pm$ 33.4 \\ \hline \hline
Bo \etal \cite{Bo-2010}*  & 38.2 $\pm$ 21.4   & 32.8 $\pm$ 23.1     & 40.2 $\pm$ 23.2     & 42.0 $\pm$ 12.9     & 34.7 $\pm$ 16.6     & 46.4 $\pm$ 28.9  & 39.1  $\pm$ 21.0 \\ \hline \hline

\multicolumn{8}{|c|}{\textbf{(a)} Our Approach (MoCap from HumanEva dataset)}                                                                                                          \\ \hline
Iteration-\rom{1}                   & 40.1 $\pm$ 34.5   &  33.1 $\pm$ 27.7   &  47.5 $\pm$ 35.2   & 48.6 $\pm$ 33.3  &  43.6 $\pm$ 31.5    &  40.0 $\pm$ 27.9          & 42.1  $\pm$ 31.6 \\ \hline
Iteration-\rom{2}                   & {\bf 35.8} $\pm$ 34.0   &  32.4 $\pm$ 26.9   &  {\bf 41.6} $\pm$ 35.4   & {\bf 46.6} $\pm$ 30.4  &  41.4 $\pm$ 31.4    &  35.4 $\pm$ 25.2          & {\bf 38.9}  $\pm$ 30.5 \\ \hline \hline

\multicolumn{8}{|c|}{\textbf{(b)} Our Approach (MoCap from CMU dataset)}                                                                                                                \\ \hline
Iteration-\rom{1}                   & 54.5 $\pm$ 23.7   &  54.2 $\pm$ 21.4   &  64.2 $\pm$ 26.7   & 76.2 $\pm$ 23.8  &  74.5 $\pm$ 19.6     & 58.3 $\pm$ 23.7           & 63.6  $\pm$ 23.1 \\ \hline
Iteration-\rom{2}                   & 52.2 $\pm$ 20.5   &  51.0 $\pm$ 15.1   &  62.8 $\pm$ 27.4   & 74.5 $\pm$ 23.2  &  72.4 $\pm$ 20.6     & 56.8 $\pm$ 21.4           & 61.6  $\pm$ 21.4 \\ \hline
\end{tabular}
}
\caption{Comparison with other state-of-the-art approaches on the HumanEva-I dataset. The average 3D pose error (mm) and standard deviation are reported for all three subjects (S1, S2, S3) and camera C1. * denotes a different evaluation protocol.
\textbf{(a)} Results of the proposed approach with one or two iterations and motion capture data from the HumanEva-I dataset. \textbf{(b)} Results with motion capture data from the CMU dataset.
}
\vspace{-3mm}
\label{tab:results}
\end{table*}
\vspace{-4mm}
\subsubsection{Comparison with State-of-the-art}\label{exp:comp}
In our experiments, we consider two sources for the motion capture data, namely HumanEva-I and the CMU motion capture dataset.
\vspace{-3.5mm}
\paragraph{HumanEva-I Dataset.}
We first use the entire 49K 3D poses of the training data as motion capture data and compare our approach with the state-of-the-art methods~\cite{Ilya_2014,Wang_2014_CVPR,SimoSerraCVPR2013,SimoSerraCVPR2012,Bo-2010,Radwan-2013iccv}. Although the training data for 2D pose estimation and 3D pose data are from the same dataset, our approach considers them as two different sources and does not know the 3D pose for a training image. We report the 3D pose error for each sequence and the average error in Table~\ref{tab:results}.
While there is no method that performs best for all sequences, our approach outperforms all other methods in terms of average 3D pose error. The approaches~\cite{Ilya_2014, Bo-2010} achieve a similar error, but they rely on stronger assumptions. In \cite{Ilya_2014} the ground-truth information is used to compute a 3D bounding volume and the inference requires around three minutes per image since the approach uses a 3D PSM.
The first iteration of our approach takes only 19 seconds per image\footnote{2D pose estimation with a pyramid of 6 scales and scale factor 0.85 (10 sec.); 3D pose retrieval (0.12 sec.); estimating projection and 2D pose refinement (7.7 sec.); 3D pose estimation (0.15 sec.); image size $640\times480$ pixels; measured on a 12-core 3.2GHz Intel processor}
and additional 8 seconds for a second iteration.

In \cite{Bo-2010} background subtraction is performed to obtain the human silhouette, which is used to obtain a tight bounding box.
The approach also uses 20 joints instead of 14, which therefore results in a different 3D pose error. We therefore use the publicly available source code~\cite{Bo-2010} and evaluate the method for 14 joints and provide the human bounding box either from ground-truth data (GT-BB) or from our 2D pose estimation (Est-BB). The results in Table~\ref{tab:results} show that the error significantly increases for \cite{Bo-2010} when the same skeleton is used and the bounding box is not given but estimated.
\vspace{-4mm}
\paragraph{CMU Motion Capture Dataset.}
In contrast to the other methods, we do not assume that the images are annotated by 3D poses but use motion capture data as a second training source. We therefore evaluate our approach using the CMU motion capture dataset~\cite{cmu_mocap} for our 3D pose retrieval. We use one third of the CMU dataset and downsample the CMU dataset from 120Hz to 30Hz, resulting in 360K 3D poses. Since the CMU skeleton differs from the HumanEva skeleton, the skeletons are mapped to the HumanEva dataset by linear regression. The results are shown in Table~\ref{tab:results}(b). As expected the error is higher due to the differences of the datasets, but the error is still low in comparison to the other methods.

\begin{figure*}[t]
\begin{center}
\includegraphics[width=.81\linewidth]{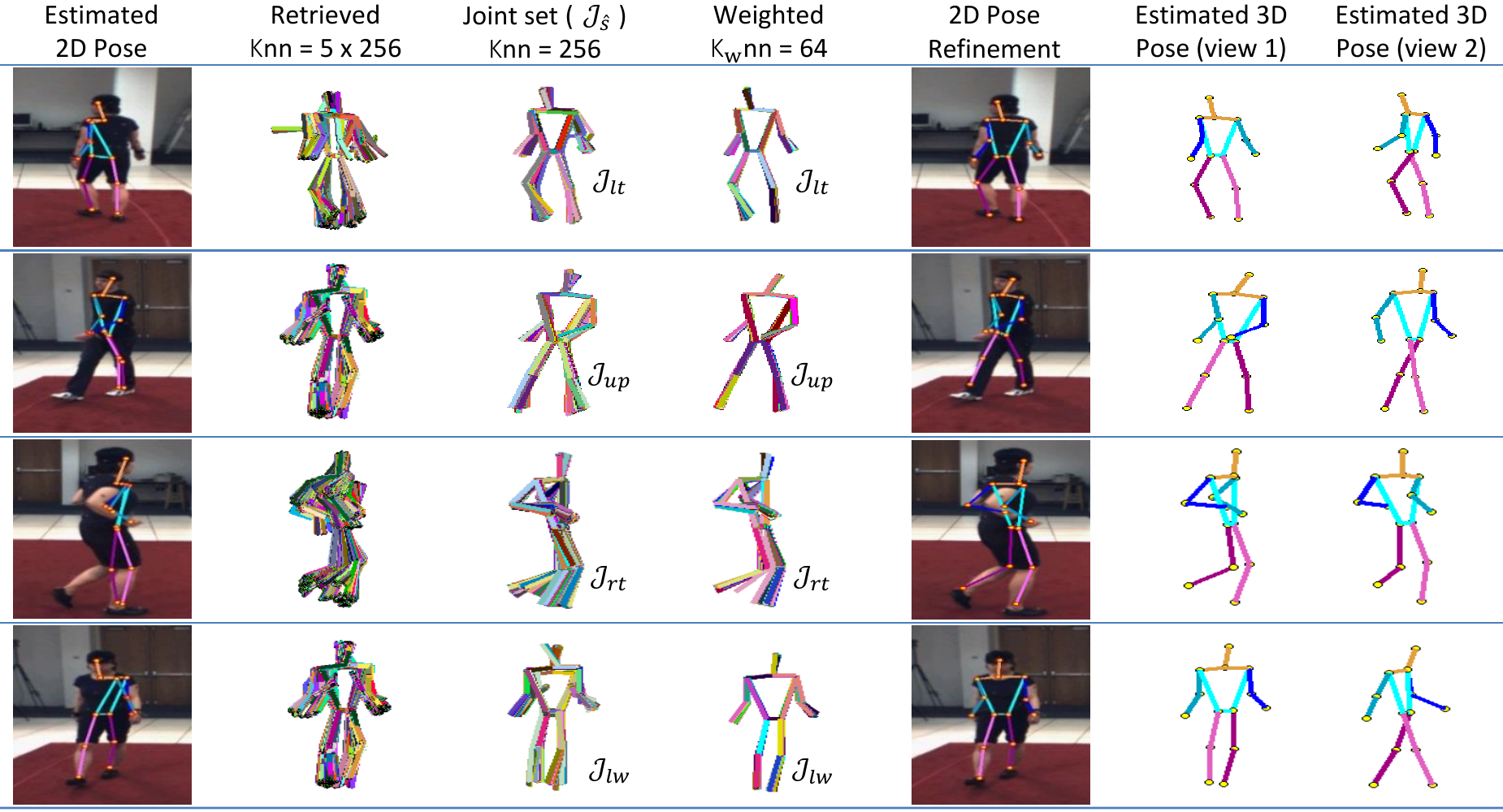}
\vspace{-2mm}
\end{center}
   \caption{
   Four examples from HumanEva-I. From left to right: estimated 2D pose ${\bf{x}}$ (Section~\ref{sec:posdet}); retrieved 3D poses from all joint sets (Section~\ref{sec:similarity}); retrieved 3D poses from inferred joint set $\mathcal{J}_{\hat{s}}$ (Section~\ref{sec:posrec}); retrieved 3D poses weighted by $w_{\mathsf{k},\hat{s}}$ \eqref{eq:wtn}; refined 2D pose $\hat{\bf{x}}$ \eqref{eqt:best_segments_pose}; estimated 3D pose $\hat{\mathbf{X}}$ \eqref{eq:energyMin2} shown from two different views.     
}
\vspace{-2mm}
\label{fig:wts2}
\end{figure*}

To analyze the impact of the motion capture data more in detail, we have evaluated the pose error for various modifications of the data in Table~\ref{tab:gt_noise}. We first remove the walking sequences from the motion capture data. The error increases for the walking sequences since the dataset does not contain poses related to walking sequences any more, but the error is still comparable with the other state-of-the-art methods (Table~\ref{tab:results}). The error for the jogging sequences actually decreases since the dataset contains less poses that are not related to jogging. In order to analyze how much of the difference between the HumanEva and the CMU motion capture data can be attributed to the skeleton, we mapped the HumanEva poses to the CMU skeletons. As shown in Table~\ref{tab:gt_noise}(c), the error increases significantly. Indeed, over 60\% of the error increase can be attributed to the difference of skeletons.
\begin{table}[t]
\centering
\scalebox{0.75}{
\begin{tabular}{|c||c|c|c||c|c|c||c|}
\hline
\multicolumn{1}{|c||}{\multirow{2}{*}{MoCap data} } & \multicolumn{3}{c||}{Walking (A1, C1)}            & \multicolumn{3}{c||}{Jogging (A2, C1)}  & \multicolumn{1}{c|}{\multirow{2}{*}{Avg.}}   \\ \cline{2-7}

\multicolumn{1}{|c||}{}                  & \multicolumn{1}{c|}{S1} & \multicolumn{1}{c|}{S2} & \multicolumn{1}{c||}{S3} & \multicolumn{1}{c|}{S1} & \multicolumn{1}{c|}{S2} & \multicolumn{1}{c||}{S3} & \\ \hline

\textbf{(a)} HuEva        & 40.1  &  33.1 &  47.5 & 48.6 &  43.6 &  40.0 & 42.1   \\ \hline
\textbf{(b)} HuEva$\setminus$Walking & 70.5 & 60.4 & 86.9 & 46.5 & 40.4 & 38.8  & 57.3\\ \hline
\textbf{(c)} HuEva-Retarget & 59.5 & 43.9 & 63.4 & 61.0 &  51.2 & 55.7 & 55.8  \\ \hline
\textbf{(d)} CMU                   & 54.5 &  54.2 &  64.2 & 76.2 &  74.5 & 58.3 & 63.6 \\ \hline
\end{tabular}
}
\caption{Impact of the MoCap data. \textbf{(a)} MoCap from HumanEva dataset. \textbf{(b)} MoCap from HumanEva dataset without walking sequences. \textbf{(c)} MoCap from HumanEva dataset but skeleton is retargeted to CMU skeleton. \textbf{(d)} MoCap from CMU dataset. The average 3D pose error (mm) is reported for the HumanEva-I dataset with one iteration.}
\label{tab:gt_noise}
\end{table}
In Table~\ref{tab:2D_results} we also compare the error of our refined 2D poses with other approaches. We report the 2D pose error for~\cite{dantone_tpami2014}, which corresponds to our initial 2D pose estimation as described in Section~\ref{sec:posdet}. In addition, we also compare our method with~\cite{YiYang-2011, Wang_CVPR2013, desai_eccv2012} using publicly available source codes. The 2D error is reduced by pose refinement using either of the two motion capture datasets and is lower than for the other methods. In addition, the error is further decreased by a second iteration. Some qualitative results are shown in Fig.~\ref{fig:wts2}.
%

\begin{table}[t]
\centering
\scalebox{0.75}{
\begin{tabular}{|c||c|c|c||c|c|c||c|}
\hline
\multicolumn{1}{|c||}{\multirow{2}{*}{Methods} } & \multicolumn{3}{c||}{Walking (A1, C1)}            & \multicolumn{3}{c||}{Jogging (A2, C1)}  & \multicolumn{1}{c|}{\multirow{2}{*}{Avg.}}   \\ \cline{2-7}

\multicolumn{1}{|c||}{}                  & \multicolumn{1}{c|}{S1} & \multicolumn{1}{c|}{S2} & \multicolumn{1}{c||}{S3} & \multicolumn{1}{c|}{S1} & \multicolumn{1}{c|}{S2} & \multicolumn{1}{c||}{S3} & \\ \hline

~\cite{dantone_tpami2014}   &  9.94 & 8.53 & 12.04 &  12.54 & 9.99 & 12.37 &                  10.90 \\
~\cite{Wang_CVPR2013}   &  17.47	 & 17.84 & 	21.24 & 	16.93 & 	15.37 & 	15.74 & 	17.43 \\
~\cite{desai_eccv2012}   &  10.44 & 9.98 & 14.47	& 14.40 & 	10.38 &	10.21 &	11.65
\\
~\cite{YiYang-2011}   &  11.83 & 10.79 & 14.28 &  14.43 & 10.49 & 11.04 & 12.14 \\
 \hline \hline

\multicolumn{8}{|c|}{\textbf{(a)} 2D Pose Refinement (with HumanEva dataset)}                                                                                                          \\ \hline
Iteration-\rom{1}        &   6.96 & 6.08 & 9.20 &  9.80 & 7.23 & 8.71 &    8.00     \\ \hline
Iteration-\rom{2}        &   {\bf 6.47} & {\bf 5.50} & {\bf 8.54} &  {\bf 9.40} & {\bf 6.79} & {\bf 7.99} &    {\bf 7.45}   \\ \hline \hline

\multicolumn{8}{|c|}{\textbf{(b)} 2D Pose Refinement (with CMU dataset)}                                                                                                                \\ \hline
Iteration-\rom{1}  & 7.62 & 6.26 & 10.99 & 11.14 & 8.58 & 9.93 &  9.08 \\ \hline
Iteration-\rom{2}  & 7.12 & 5.99 & 10.64 & 10.79 & 8.24 & 9.42 &  8.70 \\ \hline
\end{tabular}
}
\caption{2D pose estimation error (pixels) after refinement. }
\label{tab:2D_results}
\end{table}
\subsection{Evaluation on Human3.6M Dataset}
The protocol originally proposed for the Human3.6M dataset~\cite{h36m_pami} uses the annotated bounding boxes and the training data only from the action class of the test data. Since this protocol simplifies the task due to the small pose variations for a single action class and the known scale, a more realistic protocol has been proposed in~\cite{Ilya_2014} where the scale is unknown and the training data comprises all action classes. We follow the protocol~\cite{Ilya_2014} and use every $64^{th}$ frame of the subject S11 for testing.
Since the Human3.6M dataset comprises a very large number of training samples, we increased the number of regression trees for 2D pose estimation to 30 and the number of mixtures of parts to $c=40$, where each tree is trained on 10K randomly selected training images.
We use the same \textit{3D pose error} for evaluation and perform the experiments with 3D pose data from Human3.6M and the CMU motion capture dataset.
\begin{table}[t]
\centering
\scalebox{0.76}{
\begin{tabular}{|c||c||c||c|c|c|c|}
\hline
\multirow{3}{*}{Methods} & \multirow{3}{*}{~\cite{Ilya_2014}} & \multirow{3}{*}{~\cite{Bo-2010}} & \multicolumn{4}{c|}{Our Approach} \\ \cline{4-7}
                        &               &           & \multicolumn{3}{c|}{Human3.6M (Iter-I)} & \multicolumn{1}{c|}{\multirow{2}{*}{CMU (Iter-I)}}  \\ \cline{4-6}
                        &               &           & \textbf{(a)}  & \textbf{(b)} & \textbf{(c)} &  \\ \hline
3D Pose Error           & 115.7         & 117.9     & {\bf 108.3}  & 70.5 & 95.2    & 124.8              \\ \hline
\end{tabular}
}
\caption{Comparison on the Human3.6M dataset. \textbf{(a)} 2D pose estimated as in Section~\ref{sec:posdet} \textbf{(b)} 2D pose from ground-truth. \textbf{(c)} MoCap dataset includes 3D pose of subject S11.}
\label{tab:H36M_results}
\end{table}
\begin{table*}[t]
\centering
\scalebox{0.69}{
\def\arraystretch{1.1}
\begin{tabular}{|c||c|c|c|c|c|c|c|c|c|c|c|c|c|c|c|c|}
\hline
MoCap data &  Direction & Discussion & Eat &  Greet & Phone & Pose &Purchase &Sit & SitDown & Smoke &  Photo & Wait & Walk &WalkDog &WalkTogether\\ \hline\hline
H3.6M   &  88.4 & 72.5 & 108.5 &  110.2 & 97.1 &81.6 &107.2 & 119.0 &  170.8 & 108.2 & 142.5 &86.9 & 92.1& 165.7&102.0\\ \hline
H3.6M + 2D GT  &60.0&54.7&71.6&67.5&63.8&61.9&55.7&73.9&110.8&78.9&96.9& 67.9 & 47.5 &89.3 &53.4\\ \hline
H3.6M + 3D GT  &66.2&57.8&98.8&84.5&79.6&58.2&100.7&115.8&162.1&97.2&119.2&73.4&88.5&159.1&99.8\\ \hline
CMU  &102.8&80.4& 133.8&120.5& 120.7& 98.9&117.3 &150.0 & 182.6 & 135.6 &  140.1 & 104.7 & 111.3 &167.0 &116.8\\ \hline
\end{tabular}
}
\caption{The average 3D pose error (mm) on the Human3.6M dataset for all actions of subject S11.}
\vspace{-4mm}
\label{tab:H36M}
\end{table*}
\begin{figure}[t]
\begin{center}
\includegraphics[width=.871\linewidth]{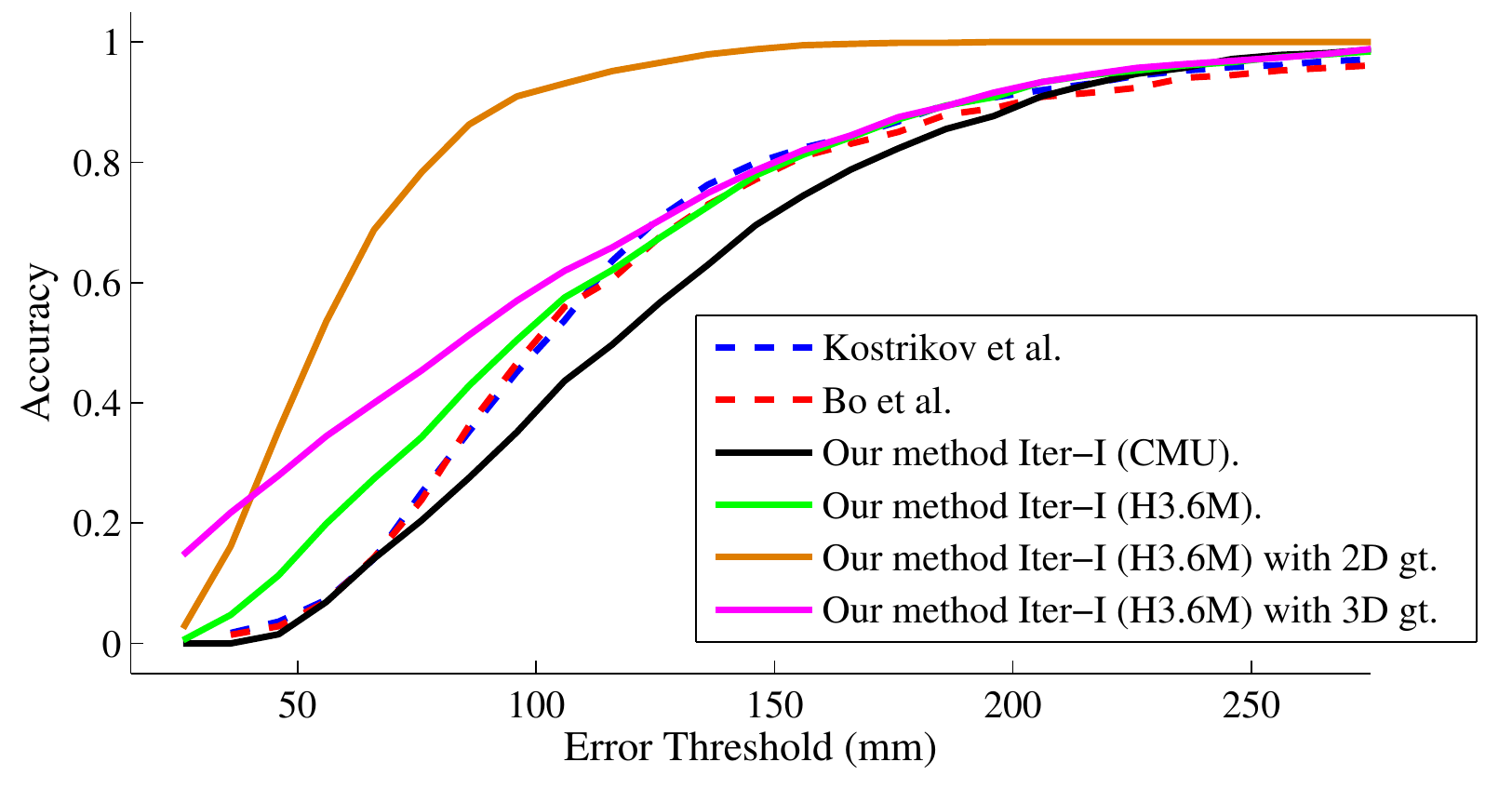}
\end{center}
   \vspace{-2mm}	
   \caption{Comparison on the Human3.6M dataset.}
\label{fig:H36M}
\end{figure}
%
In the first case, we use six subjects (S1, S5, S6, S7, S8 and S9) from Human3.6M and eliminate very similar 3D poses. We consider two poses as similar when the average Euclidean distance of the joints is less than $1.5mm$. This resulted in 380K 3D poses. In the second case, we use the CMU pose data as described in Section~\ref{exp:comp}. The results are reported in Tables~\ref{tab:H36M_results} and~\ref{tab:H36M}. Table~\ref{tab:H36M_results} shows that our approach outperforms \cite{Ilya_2014,Bo-2010}. 
On this datasets, a second iteration reduces the pose error by less than $1mm$.
Fig.~\ref{fig:H36M} provides a more detailed analysis and shows that more joints are estimated with an error below $100mm$ in comparison to the other methods. When using CMU motion capture dataset, the error is again higher due to differences of the datasets but still competitive.

We also investigated the impact of the accuracy of the initially estimated 2D poses. If we initialize the approach with the 2D ground-truth poses, the 3D pose error is drastically reduced as shown in Table~\ref{tab:H36M_results}(b) and Fig.~\ref{fig:H36M}. This indicates that the 3D pose error can be further reduced by improving the used 2D pose estimation method. In Table~\ref{tab:H36M_results}(c), we also report the error when the 3D poses of the test sequences are added to the motion capture dataset. While the error is reduced, the impact is lower compared to accurate 2D poses or differences of the skeletons (CMU). The error for each action class is given in Table~\ref{tab:H36M}.

\section{Conclusion}\label{sec:con}
In this paper, we have presented a novel dual-source approach for 3D pose estimation from a single RGB image. One source is a MoCap dataset with 3D poses and the other source are images with annotated 2D poses. In our experiments, we demonstrate that our approach achieves state-of-the-art results when the training data are from the same dataset, although our approach makes less assumptions on training and test data. Our dual-source approach also allows to use two independent sources. This makes the approach very practical since annotating images with accurate 3D poses is often infeasible while 2D pose annotations of images and motion capture data can be collected separately without much effort. 
\paragraph{Acknowledgements.} Hashim Yasin gratefully acknowledges the Higher Education Commission of Pakistan for providing the financial support. The authors would also like to acknowledge the financial support from the DFG Emmy Noether program (GA 1927/1-1) and DFG research grant (KR 4309/2-1).

{\small
\bibliographystyle{ieee}
\bibliography{egbib}
}
\end{document}